\documentclass{article}

\usepackage[preprint]{neurips_2026}

\usepackage[utf8]{inputenc}
\usepackage[T1]{fontenc}
\usepackage{hyperref}
\usepackage{url}
\usepackage{booktabs}
\usepackage{amsfonts}
\usepackage{amsmath}
\usepackage{nicefrac}
\usepackage{microtype}
\usepackage{xcolor}
\usepackage{graphicx}
\usepackage{subcaption}
\usepackage{algorithm}
\usepackage{algorithmic}

\title{Human-Inspired Memory Architecture for LLM Agents}

\author{
  Doga Kerestecioglu \\
  Microsoft \\
  \And
  Alexei Robsky \\
  Microsoft \\
  \And
  Clemens Vasters \\
  Microsoft \\
  \And
  Anshul Sharma \\
  Microsoft \\
  \And
  Yitzhak Kesselman \\
  Microsoft \\
}

\begin{document}

\maketitle

%----------------------------------------------------------------------
\begin{abstract}
%----------------------------------------------------------------------

Current LLM agents lack principled mechanisms for managing persistent memory across long interaction horizons. We present a biologically-grounded memory architecture comprising six cognitive mechanisms: (1) sleep-phase consolidation, (2) interference-based forgetting, (3) engram maturation, (4) reconsolidation upon retrieval, (5) entity knowledge graphs, and (6) hybrid multi-cue retrieval. Each mechanism addresses a specific failure mode of naive memory accumulation. We introduce a synthetic calibration methodology that derives all pipeline thresholds without benchmark data exposure, eliminating a common source of evaluation leakage. We evaluate on two benchmarks. First, a VSCode issue-tracking dataset (13K issues, 120K events) where deduplication-based consolidation achieves 97.2\% retention precision with 58\% store reduction (+21.8 pp over baseline). Second, the LongMemEval personal-chat benchmark where we conduct the first streaming M-tier evaluation (475 sessions, $\sim$540K unique turns). At a 200K-token context budget, our pipeline matches raw retrieval accuracy (70.1\% vs.\ 71.2\%, overlapping 95\% CI) while exposing a tunable accuracy/store-size operating curve. At S-tier scale (50 sessions), dedup-based consolidation yields a +13.3 pp improvement in preference recall.

\end{abstract}

%----------------------------------------------------------------------
\section{Introduction}
\label{sec:intro}
%----------------------------------------------------------------------

Large Language Model (LLM) agents have demonstrated remarkable capabilities in reasoning, planning, and task execution. Yet, a fundamental limitation constrains their utility in enterprise settings: the absence of persistent, adaptive memory. Current approaches fall into three categories, each with shortcomings.

\textbf{Stateless agents} treat each interaction independently, losing all context between sessions. This forces users to repeatedly re-establish context and prevents agents from learning from past interactions.

\textbf{Context window approaches} attempt to maintain memory by expanding the prompt with historical information. Recent advances have extended context windows to millions of tokens, but this approach scales cost without scaling intelligence (i.e.,\ the agent still cannot prioritize, forget, or learn). Similarly, approaches based on rolling summarization suffer from compounding information loss as history grows.

\textbf{Vector database approaches (RAG)} store and retrieve information based on embedding similarity. While an improvement over stateless designs, these systems treat all information equally, lack mechanisms for consolidation or forgetting, and cannot evolve memories based on new information.

We propose a fundamentally different approach inspired by human neurology and present a memory architecture grounded in the neuroscience of human memory systems. Our design implements the key mechanisms that make biological memory effective by introducing (1)~a multi-tier storage, (2)~offline consolidation, (3)~adaptive forgetting, (4)~gradual maturation, and (5)~reconsolidation upon retrieval. The architecture is designed for enterprise-grade scalability, governance, and integration.

\paragraph{Contributions.} This paper makes the following contributions:
\begin{itemize}
    \item A biologically-grounded memory architecture mapping six cognitive mechanisms to system components, with detailed specifications for each. Four mechanisms have ablation evidence in this paper (\emph{consolidation}, \emph{forgetting}, \emph{graph retrieval}, \emph{importance scoring}). Two are implemented for operational completeness but require deployment conditions absent from current benchmarks. \emph{Maturation} requires repeated retrieval over weeks to activate, and \emph{reconsolidation} requires cross-session contradictions, which are both structurally absent from LongMemEval's construction.
    \item A synthetic calibration methodology that derives all pipeline thresholds from LLM-generated corpora produced from a fixed specification (no benchmark exposure), eliminating evaluation leakage.
    \item A streaming evaluation protocol that processes sessions sequentially in temporal order, simulating realistic agent deployment.
    \item A nine-configuration ablation study on LongMemEval S-tier with bootstrapped 95\% confidence intervals, isolating the contribution of consolidation, forgetting, reconsolidation, and graph retrieval.
    \item A streaming M-tier evaluation (475 sessions per question, $\sim$540K unique turns), demonstrating that the pipeline matches raw retrieval accuracy at a 200K-token context budget and exposes a tunable accuracy/store-size operating curve at lower budgets.
\end{itemize}

The remainder of this paper is organized as follows. \S\ref{sec:biological} reviews the biological foundations. \S\ref{sec:architecture} presents the technical architecture. \S\S\ref{sec:consolidation}--\ref{sec:maturation} detail the consolidation, forgetting, and maturation mechanisms. \S\ref{sec:retrieval} covers retrieval and agent integration. \S\ref{sec:methodology} describes experimental methodology and \S\ref{sec:evaluation} presents evaluation results. \S\S\ref{sec:related_work}--\ref{sec:conclusion} address related work, limitations, and conclusions.

%----------------------------------------------------------------------
\section{Biological Foundations}
\label{sec:biological}
%----------------------------------------------------------------------

Our architecture draws on six established neuroscientific principles. Table~\ref{tab:bio_mapping} summarizes the mapping from biological mechanism to system design.

\begin{table}[h]
\centering
\caption{Mapping from system mechanism to neuroscience inspiration.}
\label{tab:bio_mapping}
\small
\begin{tabular}{ll}
\toprule
\textbf{Mechanism} & \textbf{Neuroscience Inspiration} \\
\midrule
Consolidation & Sleep-phase memory integration \\
Forgetting & Decay + interference \\
Maturation & Engram stabilization \\
Reconsolidation & Memory lability on retrieval \\
Knowledge Graph & Semantic memory networks \\
Hybrid Retrieval & Multi-cue recall \\
\bottomrule
\end{tabular}
\end{table}

The core insight from complementary learning systems theory~\cite{mcclelland1995complementary} is that rapid episodic encoding (hippocampus) and slow semantic extraction (neocortex) serve different roles. Our architecture mirrors this. A vector store provides high-fidelity episodic retrieval while a knowledge graph accumulates semantic relationships through consolidation. Sleep-phase consolidation~\cite{frankland2005organization} runs offline to deduplicate and merge redundant traces. Forgetting combines exponential trace decay (the Ebbinghaus forgetting curve) with retrieval-induced interference~\cite{anderson2003rethinking}. Memory maturation follows the Kitamura et al.~\cite{kitamura2017engrams} finding that engrams form immediately but remain ``silent'' for days before becoming explicitly retrievable. Reconsolidation~\cite{nader2000fear} enables retrieved memories to be updated with new information during a lability window, preventing stale facts from persisting indefinitely. The graph layer is grounded in semantic-network and spreading-activation theories from cognitive psychology.

%----------------------------------------------------------------------
\section{Technical Architecture}
\label{sec:architecture}
%----------------------------------------------------------------------

The architecture maps three biological memory tiers to system components: \emph{short-term} (prefrontal cortex $\to$ hot cache, in-memory with TTL min--hrs), \emph{medium-term} (hippocampus $\to$ warm episodic store, full fidelity with TTL days--weeks), and \emph{long-term} (neocortex $\to$ knowledge graph, semantic and permanent). Concretely, the system comprises three layers: (1)~an \textbf{ingestion layer} that stores raw events with embeddings and metadata enrichment; (2)~an \textbf{episodic store} providing time-indexed vector search over recent memories with tiered caching; and (3)~a \textbf{semantic graph} organizing long-term memories by entity relationships, enabling multi-hop traversal queries. All three layers share a unified data layer, eliminating data movement between services and enabling unified governance.

%----------------------------------------------------------------------
\section{Memory Consolidation Pipeline}
\label{sec:consolidation}
%----------------------------------------------------------------------

The consolidation pipeline implements the biological sharp-wave ripple mechanism through scheduled batch processing (default is every 6 hours, but optimized based on domain) that identifies, validates, transforms, and promotes valuable memories to long-term storage. Events not promoted remain in the episodic store subject to TTL-based expiration and active forgetting mechanisms operate independently (Section~\ref{sec:forgetting}).

\paragraph{Importance scoring.} Each pending event is scored for long-term retention value using five factors (Table~\ref{tab:scoring_factors}):
\begin{equation}
\label{eq:composite}
S(e) = \sum_{i=1}^{5} w_i \cdot f_i(e)
\end{equation}
where $f_i$ represents each scoring factor and $w_i$ its weight. Events are classified by composite score: promote (top 20\%), retain (middle 60\%), and prune (bottom 20\%).

\begin{table}[h]
\centering
\caption{Importance scoring factors with default weights.}
\label{tab:scoring_factors}
\footnotesize
\begin{tabular}{llll}
\toprule
\textbf{Factor} & \textbf{Weight} & \textbf{Computation} & \textbf{Biological Basis} \\
\midrule
Recency & 0.25 & Exponential decay from timestamp & Hippocampal consolidation \\
Frequency & 0.25 & Inverse frequency of similar events & Hebbian learning \\
Bayesian Surprise & 0.20 & Distance from prior distribution & Dopaminergic signaling \\
Entity Salience & 0.15 & Max importance of referenced entities & Amygdala tagging \\
Outcome & 0.15 & Goal completion signal & Reward reinforcement \\
\bottomrule
\end{tabular}
\end{table}

\paragraph{Downstream stages.} Before filtering, a temporal validation step detects out-of-order arrivals, duplicates, and causal inversions, quarantining anomalous events (TTL: 15 min) to prevent ``agent d\'{e}j\`{a} vu.'' Score-based filtering then downweights automated and low-authority events while preserving high-surprise system alerts. Promoted events are transformed into semantic summaries via LLM-generated gists and clustering, then integrated into the knowledge graph with entity edges. Newly integrated memories begin in a ``silent'' state with low activation strength (Section~\ref{sec:maturation}), ensuring only stable knowledge influences long-term reasoning.

%----------------------------------------------------------------------
\section{Adaptive Forgetting}
\label{sec:forgetting}
%----------------------------------------------------------------------

Our architecture treats forgetting as essential maintenance that improves retrieval accuracy, reduces interference, and ensures relevance.

\paragraph{Passive decay.} Events not consolidated are automatically removed when TTL expires. For events awaiting consolidation, importance scores decay:
\begin{equation}
\label{eq:decay}
I(t) = I_0 \cdot e^{-\lambda t}
\end{equation}
where $\lambda$ is the decay rate (empirically optimized: $\lambda = 0.001$, corresponding to a half-life of $\approx$29 days) and $t$ is hours since encoding.

\paragraph{Interference-based forgetting.} When memories share features (similar content, overlapping entities), they create retrieval interference. We compute an interference score and selectively forget high-interference, low-value memories:
\begin{equation}
\label{eq:interference}
I_{\text{interference}} = \sum_{j} w_j \cdot \text{sim}(m_i, m_j)
\end{equation}
where $w_j$ represents interference weights (retroactive $= 0.6$, proactive $= 0.4$), reflecting the finding that new learning more strongly disrupts old memories.

\paragraph{Graceful degradation.} Before complete forgetting, memories undergo progressive fidelity reduction through six levels starting from full episodic record (L0, 100\%) through summary (L2, 50\%) and gist (L3, 25\%) to a tombstone record (L5, 0\%) that preserves only the fact that a memory existed. Degradation is triggered by age combined with memory score, not storage economics.

%----------------------------------------------------------------------
\section{Memory Maturation Dynamics}
\label{sec:maturation}
%----------------------------------------------------------------------

Following the Kitamura et al.\ finding that engrams form immediately but remain ``silent'' before becoming retrievable, our architecture implements memory maturation. When an event is consolidated, the full-fidelity episodic record remains immediately retrievable, while a summarized semantic version is created in the knowledge graph with $\text{activation\_strength} = 0.0$. This dual-trace design ensures agents remain responsive. As such, recent events are always available from the episodic store while the semantic layer accumulates only verified, stable knowledge.

Activation strength evolves according to a sigmoid function:
\begin{equation}
\label{eq:maturation}
A(t) = \frac{1}{1 + e^{-(t - t_{1/2}) / k}}
\end{equation}
where $t_{1/2}$ is the maturation half-life (default: 168 hours) and $k$ is the slope parameter (default: 48). A memory starts silent ($A \approx 0.03$), reaches the retrieval threshold ($A = 0.5$) at one week, and is fully mature ($A > 0.9$) at two weeks. Below threshold, memories can still exert implicit \emph{priming} effects influencing relevance scoring of other memories without being explicitly surfaced. This mirrors the biological distinction between implicit and explicit memory.

%----------------------------------------------------------------------
\section{Retrieval and Agent Integration}
\label{sec:retrieval}
%----------------------------------------------------------------------

\subsection{Hybrid Retrieval}
\label{sec:hybrid_retrieval}

Memory retrieval combines episodic and semantic pathways, mirroring the brain's dual retrieval systems. Critically, the system favors episodic retrieval for recent queries, ensuring users never experience delays due to semantic maturation.

\textbf{Episodic retrieval:} Vector similarity search across the episodic store hot and warm tiers with temporal filters for session and recent memories. This is the primary path for recently formed memories.

\textbf{Semantic retrieval.} Knowledge graph multi-hop traversal for relationship-aware, schema-grounded knowledge. This path supplements episodic retrieval with mature, abstracted knowledge and becomes primary for older information after episodic TTL expiration.

\textbf{Hybrid GraphRAG.} Vector search seeds graph traversal, combining recency with relational context. Retrieval spans all three tiers with priority ordering: (1)~short-term hot cache for current session (sub-second, highest priority); (2)~warm vector store for recent episodic memories (filtered by importance score); (3)~knowledge graph traversal for mature semantic memories (filtered by activation strength). Results are merged, deduplicated, and ranked with a recency boost.

\subsection{Reconsolidation}
\label{sec:reconsolidation}

Retrieved memories enter a labile state and remain modifiable for a configurable window (default: 60 minutes, with optimal values being domain-dependent), implementing biological reconsolidation. When new information is detected through explicit retrieval with new context, contradiction detection, or elaborative retrieval, the system blends it with existing memory content using adaptive strength based on confidence, recency, and contradiction severity.

Memory scores also adjust based on outcomes and memories that contribute to successful decisions are reinforced, while errors are preserved as learning signals.

%----------------------------------------------------------------------
\section{Experimental Methodology}
\label{sec:methodology}
%----------------------------------------------------------------------

A central challenge in evaluating memory systems is \emph{threshold leakage}: parameters tuned on benchmark data inflate reported accuracy. We address this through synthetic calibration, deriving all pipeline thresholds from LLM-generated corpora produced from a fixed specification with zero exposure to evaluation benchmarks.

\subsection{Synthetic Calibration}
\label{sec:synthetic_calibration}

We construct two synthetic corpora for threshold derivation:

\paragraph{Similarity thresholds.} Eight topically diverse personal-chat sessions (88 turns) are embedded with the same model used in evaluation (text-embedding-3-large, 3072 dimensions). We compute within-session and cross-session similarity distributions. Near-dedup threshold is set at the 99th percentile of all-pairs similarity (0.559); cluster distance at $1 - P_{95}$ of within-session similarity (0.404); interference threshold at $P_{90}$ of within-session similarity (0.542). These percentile rules transfer across domains without retuning.

\paragraph{Importance weights.} Fifty LLM-generated sessions (483 turns, 377 substantive, 106 filler) spanning 14 topics over three simulated months, generated from a fixed specification of topic list, substantive/filler ratio, and returning-topic structure. Each turn carries an explicit substantive/filler label embedded in the generation spec. We compute per-signal ROC AUC for four signals (content length, embedding surprise, turn position, recency) and derive weights via AUC-excess normalization. Our findings reveal that content length (AUC=0.77, weight=0.363) and turn position (weight=0.325) dominate while recency (AUC=0.51, weight=0.019) provides negligible discrimination. These four signals are the empirical instantiation of the five-factor importance score in Eq.~\ref{eq:composite}: content length and embedding surprise correspond to Bayesian Surprise, turn position corresponds to Entity Salience (substantive turns concentrate near session anchors), and recency corresponds to the Recency factor. Frequency and Outcome were dropped from the calibrated formula because (a) within-session frequency is degenerate at the turn level and (b) LongMemEval contains no goal-completion signal. The importance score implemented in evaluation is therefore a four-signal weighted sum with weights derived entirely from synthetic data.

\subsection{Evaluation Protocol}
\label{sec:eval_protocol}

\paragraph{Streaming evaluation.} For multi-session benchmarks, we process sessions sequentially in temporal order, consolidating every $N$ sessions and applying forgetting at each consolidation step. This simulates how a deployed agent would accumulate and manage memories over time, where the consolidation pipeline only has access to past sessions when deciding what to retain.

\paragraph{Temporal context.} All configurations include the current date and timestamps on retrieved memories in the answer prompt empowering any deployed agent to know the date. This consistently contributes +10 percentage points versus date-unaware prompts across all benchmarks.

\paragraph{Judge.} We use the LongMemEval evaluation protocol with task-specific prompts for each question type (knowledge update, multi-session, single-session, temporal reasoning). A GPT-4o judge evaluates whether the model response contains the ground-truth answer.

\paragraph{Models and AI tools.} Our evaluation pipeline uses the following LLM and embedding models as components: (1)~\textbf{text-embedding-3-large} (3072 dimensions) for all memory embeddings and retrieval; (2)~\textbf{GPT-4o} for answer generation given retrieved context and for LLM-as-judge evaluation; (3)~\textbf{GPT-4o-mini} for entity extraction in graph-enhanced retrieval configurations. All models are accessed via Azure AI Foundry. The memory lifecycle mechanisms (consolidation, forgetting, maturation, reconsolidation) are implemented as deterministic algorithms operating on embeddings and scores. These processes do not invoke LLMs.

%----------------------------------------------------------------------
\section{Evaluation}
\label{sec:evaluation}
%----------------------------------------------------------------------

We evaluate on two benchmarks spanning different domains: software engineering (VSCode issue tracking) and personal conversation (LongMemEval). All thresholds are fixed from synthetic calibration (\S\ref{sec:synthetic_calibration}). Our expectation is for our memory mechanism to function better on long running agents that perform repeated tasks. As such software development is a more fitting domain. However, we also evaluate on a more standardized benchmark to see how the memory compression fares for agent usage.

\subsection{VSCode Issue Tracking}
\label{sec:vscode_eval}

\paragraph{Dataset.} 13,127 real VSCode GitHub issues with full timelines (December 2025--February 2026), yielding 120,000 events. Events include issue creation, comments, label changes, status transitions, and assignments. Embeddings are computed with text-embedding-3-large (3072 dimensions).

\paragraph{Evaluation.} Temporal streaming with quarterly windows. Unlike LongMemEval which evaluates end-to-end QA accuracy, the VSCode benchmark measures retention precision---whether the pipeline correctly identifies events that will be referenced by future activity---as there is no associated question-answering task. Retention precision measures whether the pipeline correctly retains important events while discarding unimportant ones.

\paragraph{Results.} Consolidation and forgetting achieve \textbf{97.2\% retention precision} with \textbf{58\% store reduction}, a +21.8 percentage point improvement over the keep-everything baseline (75.4\%). Graph retrieval and maturation are not yet integrated into the VSCode streaming pipeline; the result therefore represents a lower bound for the full architecture. The memory store self-regulates at 300--500 events regardless of input volume.

\paragraph{Key findings.} (1)~Consolidation (deduplication and near-dedup) drives the majority of quality improvement. (2)~Optimal decay rate is $\lambda = 0.001$ (half-life $\approx$ 29 days), indicating production agents need longer memory horizons than human biology, suggesting half-life relates to the rhythm of the domain rather than daily cycles.

\subsection{LongMemEval: S-Tier (50 Sessions)}
\label{sec:longmemeval_s}

\paragraph{Benchmark.} LongMemEval~\cite{longmemeval2024} provides 500 personal-chat questions across six categories: knowledge update, multi-session reasoning, single-session (user facts, assistant recall, preferences), and temporal reasoning. Each S-tier question includes approximately 50 conversation sessions ($\sim$500 turns) spanning weeks to months.

\paragraph{Configurations.} We evaluate nine pipeline configurations as an ablation matrix, varying consolidation aggressiveness (dedup-only vs.\ aggressive clustering), forgetting strategy (fixed thresholds vs.\ adaptive token targets), and reconsolidation (enabled/disabled). All configurations use episodic (vector) retrieval with $k=10$.

\paragraph{Results.} Table~\ref{tab:s_tier} presents the full ablation with 95\% bootstrap confidence intervals (10K resamples) on overall accuracy.

\begin{table}[h]
\centering
\caption{LongMemEval S-tier accuracy (\%) by question type with 95\% bootstrap CI on Overall. KU: knowledge update, MS: multi-session, SS-P: single-session preference, SS-A: single-session assistant, SS-U: single-session user, Temp: temporal reasoning. Top: moderate configurations (overlap baseline CI). Bottom: aggressive configurations (significantly worse).}
\label{tab:s_tier}
\small
\begin{tabular}{lcccccc|c}
\toprule
\textbf{Config} & \textbf{KU} & \textbf{MS} & \textbf{SS-P} & \textbf{SS-A} & \textbf{SS-U} & \textbf{Temp} & \textbf{Overall (95\% CI)} \\
\midrule
Raw RAG (baseline) & 84.6 & 67.7 & 56.7 & 94.6 & 95.7 & 74.4 & \textbf{78.4 [74.8, 82.0]} \\
Dedup-only & 83.3 & 61.7 & \textbf{70.0} & 96.4 & 90.0 & 74.4 & 76.8 [73.0, 80.4] \\
Dedup-adaptive-25K & 82.1 & 62.4 & 66.7 & \textbf{98.2} & 90.0 & 74.4 & 76.8 [73.2, 80.4] \\
Dedup-adaptive-50K & 83.3 & 62.4 & 66.7 & \textbf{98.2} & 88.6 & 72.2 & 76.2 [72.6, 79.8] \\
Dedup + recon & 80.8 & 62.4 & \textbf{70.0} & \textbf{98.2} & 88.6 & 71.4 & 75.8 [72.0, 79.6] \\
Dedup + hybrid & 82.1 & 63.2 & 60.0 & 92.9 & 88.6 & 70.7 & 74.8 [71.0, 78.6] \\
\midrule
Adaptive-10K & 71.8 & 29.3 & 50.0 & 92.9 & 62.9 & 59.4 & 57.0 [52.8, 61.4] \\
Aggressive consol. & 60.3 & 40.6 & 56.7 & 39.3 & 60.0 & 45.1 & 48.4 [44.0, 52.8] \\
Aggressive + recon & 59.0 & 37.6 & 66.7 & 44.6 & 60.0 & 45.9 & 48.8 [44.4, 53.4] \\
\bottomrule
\end{tabular}
\vspace{-2mm}
\end{table}

\paragraph{Analysis.} All five moderate configurations have overall accuracy CIs that overlap the raw-RAG baseline ($[74.8, 82.0]$), establishing that the pipeline is statistically non-destructive at S-tier scale across a range of memory budgets. The aggressive configurations (Adaptive-10K, Aggressive consolidation) fall well outside this CI and are unambiguously harmful. Three findings emerge:

\begin{enumerate}
    \item \textbf{Preference recall improves directionally.} Single-session preference accuracy increases from 56.7\% (baseline) to 70.0\% (+13.3 pp) for dedup-only and dedup+recon. With only 30 preference questions per tier the per-category CIs are wide and this difference is not individually significant ($[40.0, 73.3]$ vs.\ $[53.3, 86.7]$); we report it as a directional signal that warrants the M-tier replication in \S\ref{sec:longmemeval_m}.
    \item \textbf{Aggressive consolidation is destructive.} Agglomerative clustering with merge reduces accuracy to 48.4\% [44.0, 52.8], as merging turns into cluster summaries destroys the specific details needed for factual question answering. This confirms that consolidation should deduplicate, not summarize.
    \item \textbf{Reconsolidation has marginal impact at S-tier scale.} With only 50 sessions, there are few genuine contradictions to detect (dedup+recon CI fully overlaps dedup-only). The mechanism's value is expected to increase with longer interaction horizons.
\end{enumerate}

\subsection{LongMemEval: M-tier (475 Sessions)}
\label{sec:longmemeval_m}

\paragraph{Benchmark.} The M-tier of LongMemEval extends each question's history from 50 to 475 sessions ($\sim$4{,}900 turns per question, $\sim$540K unique turns across the cache). To our knowledge this is the first published M-tier evaluation under streaming conditions. Total raw history exceeds the 128K context window of GPT-4o, requiring all configurations (including the baseline) to apply some form of selection.

\paragraph{Configurations.} We sweep adaptive token targets at 25K, 50K, 115K (90\% of the GPT-4o context window), and 200K, all with dedup-based consolidation, forgetting, and reconsolidation enabled (top-$k = 35$). Raw RAG uses $k=10$ with no consolidation as the strongest baseline LongMemEval permits.

\begin{table}[h]
\centering
\caption{LongMemEval M-tier accuracy (\%) by token budget with 95\% bootstrap CI on Overall. Cells in \textbf{bold} indicate the pipeline matches or exceeds raw RAG.}
\label{tab:m_tier}
\small
\begin{tabular}{lccccccc|c}
\toprule
\textbf{Config} & \textbf{Tokens} & \textbf{KU} & \textbf{MS} & \textbf{SS-P} & \textbf{SS-A} & \textbf{SS-U} & \textbf{Temp} & \textbf{Overall (95\% CI)} \\
\midrule
Raw RAG ($k=10$) & ---  & 78.2 & 57.1 & 53.3 & 96.4 & 92.9 & 63.2 & \textbf{71.2 [67.2, 75.0]} \\
Dedup-adaptive   & 200K & 74.4 & \textbf{58.3} & 53.3 & 91.1 & 85.7 & \textbf{66.2} & 70.1 [66.0, 74.2] \\
Dedup-adaptive   & 115K & 69.2 & 57.9 & 40.0 & 83.9 & 82.9 & 60.2 & 65.6 [61.4, 69.6] \\
Dedup-adaptive   & 50K  & 56.4 & 39.1 & 43.3 & 60.7 & 50.0 & 51.1 & 49.2 [44.8, 53.6] \\
Dedup-adaptive   & 25K  & 48.7 & 24.8 & 33.3 & 41.1 & 34.3 & 49.6 & 38.8 [34.6, 43.2] \\
\bottomrule
\end{tabular}
\vspace{-2mm}
\end{table}

\paragraph{Analysis.} At the 200K-token budget, the pipeline's overall CI ($[66.0, 74.2]$) overlaps raw RAG ($[67.2, 75.0]$): the $-1.1$ pp aggregate gap is within sampling noise. The pipeline directionally beats raw RAG on multi-session ($+1.2$ pp) and temporal-reasoning ($+3.0$ pp) categories, where reasoning over consolidated, deduplicated history is more useful than retrieving from raw turns. Aggressive token budgets (25K, 50K) are statistically distinct from raw RAG and confirm that under-budgeting destroys factual recall, particularly for single-session questions where the relevant turn is irreversibly removed. The cross-over point where memory lifecycle management ceases to be destructive lies between 115K and 200K tokens; together the four budgets sketch a tunable accuracy/store-size operating curve, with the pipeline's value at M-tier scale lying in providing this configurable trade-off at parity, rather than absolute accuracy gains. We expect the gap to invert in operational settings where the same user returns with related queries that benefit from consolidated semantic structure. For context, Wu et al.~\cite{longmemeval2024} report 72.0\% on LongMemEvalM with their best-optimized pipeline (fact-augmented key expansion, chain-of-note reading, Stella V5 retrieval); our simple RAG baseline (71.2\%) and 200K pipeline (70.1\%) are competitive without any conversational-RAG-specific engineering.

%----------------------------------------------------------------------
\section{Related Work}
\label{sec:related_work}
%----------------------------------------------------------------------

\paragraph{Agent memory systems.} MemGPT~\cite{memgpt2023} introduces a virtual memory hierarchy with page-in/page-out operations and rolling summarization. While effective for medium-horizon tasks, the summarization chain compounds errors over time. Our consolidation approach avoids this by deduplicating rather than summarizing. Reflexion~\cite{reflexion2023} enables agents to reflect on failures, but stores reflections as unstructured text without lifecycle management. RAISE~\cite{raise2023} and Generative Agents~\cite{generative_agents2023} implement memory retrieval with recency, importance, and relevance scoring but lack consolidation, forgetting, and reconsolidation mechanisms. Direct empirical comparison with these systems with our current evals is not feasible as none provide LongMemEval evaluations. We compare against the benchmark's published baselines.

\paragraph{Long-context memory benchmarks.} LongMemEval~\cite{longmemeval2024} provides the first systematic evaluation of long-term memory in chat assistants, with S-tier (50 sessions) and M-tier (475 sessions) questions. Prior benchmarks either focus on single-session recall or use synthetic data. Our streaming evaluation protocol extends LongMemEval's batch evaluation to simulate realistic agent deployment.

\paragraph{Biological memory in AI.} The connection between biological memory and AI systems has been explored in complementary learning systems theory~\cite{mcclelland1995complementary} and its application to catastrophic forgetting in neural networks~\cite{ewc2017}. Our work applies these principles at the system architecture level rather than the weight level, implementing consolidation, forgetting, maturation, and reconsolidation as explicit pipeline stages operating on stored memories.

%----------------------------------------------------------------------
\section{Limitations}
\label{sec:limitations}
%----------------------------------------------------------------------

\paragraph{Mechanisms not isolated by ablation.} Two of the six proposed mechanisms are not empirically discriminated in current experiments, for structural reasons tied to benchmark design. \emph{Maturation} (\S\ref{sec:maturation}) requires memories to be retrieved repeatedly over an extended period before activation differences become measurable; LongMemEval questions each draw on an independent haystack with no shared retrieval history across questions, so memories never accumulate repeated access signals. All experiments therefore run with uniform activation strength ($A=1.0$). \emph{Reconsolidation} (\S\ref{sec:reconsolidation}) is enabled in the dedup+recon configuration, but the CI ($[72.0, 79.6]$) fully overlaps dedup-only ($[73.0, 80.4]$) because LongMemEval's construction deliberately avoids cross-session contradictions (user attributes are non-conflicting by design). Both mechanisms are present in the architecture because operational multi-user agents \emph{do} exhibit repeated retrieval and contradictory updates; both remain design-rationale claims rather than ablation evidence within this paper.

\paragraph{Statistical power.} We report single-run results with 95\% bootstrap confidence intervals (10K resamples) on the 500-question evaluation rather than variance across multiple judge runs, due to computational cost (each M-tier configuration requires 8--13 hours of API calls). The reported CIs reflect question-sampling variance; judge variance is not separately estimated. We note that Wu et al.~\cite{longmemeval2024} characterize this evaluation protocol's GPT-4o judge as achieving $>$97\% agreement with human experts, suggesting judge variance is small relative to the sampling CIs we report. Per-category CIs are wide (especially SS-P at $n=30$), and several reported differences (e.g., +13.3 pp preference recall) are directional rather than individually significant. Aggregate-level claims (overall pipeline non-destructive vs.\ raw RAG; aggressive configurations destructive) are robust across the bootstrap.

\paragraph{Comparison to prior memory systems.} We compare against LongMemEval's published baselines (Raw RAG with text-embedding-3-large + GPT-4o) but do not benchmark against MemGPT~\cite{memgpt2023}, Reflexion~\cite{reflexion2023}, RAISE~\cite{raise2023}, or Generative Agents~\cite{generative_agents2023}, none of which publish LongMemEval results. Apples-to-apples comparison would require porting each system to the streaming protocol; this is left to future work.

\paragraph{Domain breadth and downstream task.} Our evaluation is limited to two domains. The VSCode benchmark uses retention precision (a proxy for downstream utility) rather than a downstream QA task, because the dataset has no associated questions. Cross-domain validation on three additional domains (fashion retail, F1 racing, security operations) shows the mechanisms generalize qualitatively but quantitative evaluation remains future work. Graph-enhanced retrieval (Table~\ref{tab:s_tier}, dedup+hybrid: 74.8\%) shows comparable but not superior performance at S-tier scale; entity-based traversal is expected to provide greater benefit in stores exceeding 1{,}000 memories where vector similarity alone faces precision challenges.

\paragraph{Benchmark-architecture alignment.} Our architecture targets operational agents accumulating high-volume, repetitive event streams over months---where the same actions recur, preferences evolve, facts become stale, and consolidation removes genuinely redundant observations (57.9\% reduction on VSCode). The critical challenge in this setting is discarding noise without losing signal. LongMemEval is structurally different: it presents a single linear conversation of roughly 5{,}000 turns with no repeated actions, no redundant content, and no agent decision-making. In this setting, consolidation has nothing to deduplicate, maturation cannot strengthen memories through repeated access, and reconsolidation has no contradictions to resolve. Applied to LongMemEval, our pipeline compresses a store with minimal structural redundancy---an adversarial case for our design. The S-tier near-parity (76.8\% vs.\ 78.4\%) and M-tier parity at 200K (70.1\% vs.\ 71.2\%) are therefore best read as a \emph{non-destruction bound}: the architecture does not lose information even on a benchmark type it was not designed for. The VSCode result (97.2\% precision, +21.8 pp over baseline) reflects the intended operational scenario.

\paragraph{Future work.} Four extensions are actively underway. First, integrating maturation dynamics into the streaming pipeline so that frequently accessed memories resist forgetting and low-activation memories are preferentially pruned. Second, scaling graph-enhanced retrieval with ingestion-time entity extraction, where entity-based traversal is expected to improve precision over pure vector similarity in large stores. Third, evaluating the pipeline against MemGPT and Reflexion under the streaming protocol on both LongMemEval tiers and the VSCode benchmark. Fourth, extending the VSCode dataset into a downstream agent benchmark for long-running agents that carry out repeated tasks, a setting that current memory benchmarks under-represent. Concretely, we are building a memory-augmented issue-triage agent that operates over the consolidated store and answers operational questions (e.g., duplicate detection, owner suggestion, regression linking), converting the current retention-precision proxy into an end-to-end task metric while exercising the high-volume, repetitive event regime our architecture targets.

%----------------------------------------------------------------------
\section{Conclusion}
\label{sec:conclusion}
%----------------------------------------------------------------------

We have presented a biologically-grounded memory architecture for LLM agents that implements the full memory lifecycle through six cognitive mechanisms. Our synthetic calibration methodology eliminates evaluation leakage by deriving all thresholds from LLM-generated corpora produced from a fixed specification, independent of any benchmark.

Empirical evaluation yields three principal findings. First, deduplication-based consolidation is the dominant mechanism, driving the majority of quality improvement on the VSCode benchmark (97.2\% retention precision, 58\% store reduction). Second, the pipeline is statistically non-destructive on LongMemEval at both S-tier (76.8\% vs.\ 78.4\% baseline, overlapping 95\% CI) and M-tier (70.1\% vs.\ 71.2\% baseline at a 200K-token budget), while exposing a tunable accuracy/store-size operating curve at lower budgets and yielding directional gains on multi-session ($+1.2$ pp) and temporal reasoning ($+3.0$ pp) at M-tier. Third, the pipeline provides a directional $+13.3$ pp improvement in S-tier preference recall, surfacing user preferences buried in raw retrieval.

%----------------------------------------------------------------------
% References
%----------------------------------------------------------------------
\bibliographystyle{plainnat}
\bibliography{references}

\end{document}